# Advancing Autonomous Vehicle Intelligence: Deep Learning and Multimodal LLM for Traffic Sign Recognition and Robust Lane Detection

Chandan Kumar Sah[1*†]
sahchandan98@buaa.edu.cn

Ankit Kumar Shaw[2*]
shawak10@mails.tsinghua.edu.cn

Xiaoli Lian[1*]
lianxiaoli@buaa.edu.cn

Arsalan Shahid Baig[1*]
arsalan@buaa.edu.cn

Tuopu Wen[2*]
wtp18@tsinghua.org.cn

Kun Jiang[2†]
jiangkun@tsinghua.edu.cn

Mengmeng Yang[2*]
yangmm_qh@mail.tsinghua.edu.cn

Diange Yang[2†]
ydg@mail.tsinghua.edu.cn

## Abstract

*Autonomous vehicles (AVs) require reliable traffic sign recognition and robust lane detection capabilities to ensure safe navigation in complex and dynamic environments. This paper introduces an integrated approach combining advanced deep learning techniques and Multimodal Large Language Models (MLLMs) for comprehensive road perception. For traffic sign recognition, we systematically evaluate ResNet-50, YOLOv8, and RT-DETR, achieving state-of-the-art performance of 99.8% with ResNet-50, 98.0% accuracy with YOLOv8, and achieved 96.6% accuracy in RT-DETR despite its higher computational complexity. For lane detection, we propose a CNN-based segmentation method enhanced by polynomial curve fitting, which delivers high accuracy under favorable conditions. Furthermore, we introduce a lightweight, Multimodal, LLM-based framework that directly undergoes instruction tuning using small yet diverse datasets, eliminating the need for initial pretraining. This framework effectively handles various lane types, complex intersections, and merging zones, significantly enhancing lane detection reliability by reasoning under adverse conditions. Despite constraints in available training resources, our multimodal approach demonstrates advanced reasoning capabilities, achieving a Frame Overall Accuracy (FRM) of 53.87%, a Question Overall Accuracy (QNS) of 82.83%, lane detection accuracies of 99.6% in clear conditions and 93.0% at night, and robust performance in reasoning about lane invisibility due to rain (88.4%) or road degradation (95.6%). The proposed comprehensive framework markedly enhances AV perception reliability, thus contributing significantly to safer autonomous driving across diverse and challenging road scenarios.*

[1]School of Computer Science and Engineering, Beihang University
[2]School of Vehicle and Mobility, Tsinghua University

## 1. Introduction

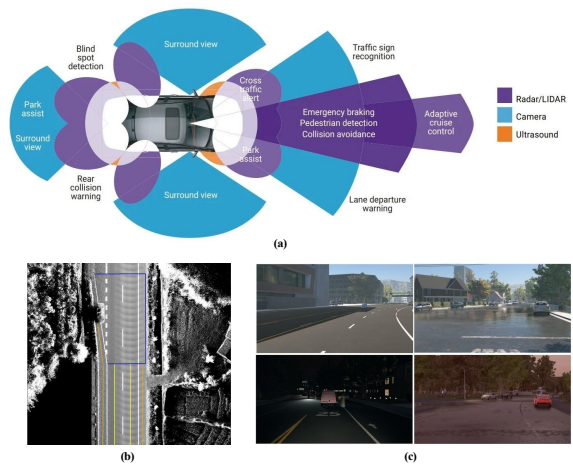

Figure 1. Multimodal LLM-Based Road Elements Understanding. (a) An AV uses multiple sensors (camera, LiDAR, radar, ultrasonic) for perception and navigation. (b) Annotated MapLM dataset with colored poly-lines: Yellow = motorway lanes, Orange = bicycle lanes, Blue = intersections/lane-change zones. (c) Apollo synthetic data representing varied times of day, weather, and road conditions.

Autonomous vehicles (AVs) rely on accurate detection of traffic signs and lane markings for safe navigation [42, 46, 54]. Errors in perception can lead to accidents, as human error accounts for over 90Deep learning has significantly advanced AV perception, with CNNs achieving high accuracy in traffic sign recognition . One-stage detectors like YOLO enable real-time sign detection [35], while transformer-based models enhance contextual reasoning [33]. Hybrid architectures, such as Local Vision Trans-former (Local-VT), outperform traditional CNNs and ViTs, achieving 99.7% accuracy on GTSRB [10]. Similarly, fine-tuned YOLOv8 reaches 97% mAP in traffic sign detection under varied lighting. Lane detection remains challenging in adverse conditions due to faded or obstructed markings. While deep networks outperform classical methods, rain and glare still degrade accuracy [33]. Recent CNN-based lane detectors trained on wet-road datasets improve robustness [12], while LVLane enhances detection under extreme lighting and occlusions.

*Equal Contribution, †Corresponding Author

This paper evaluates ResNet-50, YOLOv8, and RT-DETR for traffic sign recognition across diverse scenarios [17]. Additionally, we introduce a Multimodal LLM-based road element detection pipeline integrating vision and textual cues for enhanced lane recognition [39]. Multimodal Large Language Models (MLLMs) fuse image, LiDAR, and textual data for improved AV reasoning, while LiDAR-LLM extends this capability by aligning sparse 3D point clouds with language models for spatial understanding [52]. Our approach refines lane visibility analysis and validates map data, improving high-definition (HD) map reliability and AV safety [53]. Further, we hope to advance autonomous Vehicle research with the following contributions:

- Systematic Comparison of Traffic Sign Recognition Models: We conduct a comprehensive evaluation of ResNet-50, YOLOv8, and RT-DETR architectures for traffic sign detection, delineating their trade-offs in classification accuracy, real-time performance, and robustness to environmental challenges. This analysis provides actionable insights for deploying optimal models in practical AV systems.
- Adaptive Multimodal LLM Framework for Robust Lane Detection: We propose a multimodal LLM-augmented framework designed to enhance adaptive lane detection through contextual reasoning capabilities of Large Language Models (LLMs). The framework effectively addresses complex challenges in dynamic road environments, including diverse lane types, complex intersections, degraded markings, and occlusions, overcoming traditional methods' reliance solely on visual cues. Our approach enables the LLM to infer lane continuity and road structure even when visual information is incomplete or obscured. Additionally, task-specific prompts and annotated color-coded training data significantly boost perception accuracy and adaptability in adverse conditions such as poor weather and low visibility, thereby enhancing perception reliability and ensuring safe navigation for autonomous vehicle systems.

The remainder of the paper is organized as follows: Section 2 reviews traffic sign and lane detection research; Section 3 details model adaptations, dataset preparation, and our multimodal LLM approach; Section 4 compares model performance and lane detection improvements; and Section 5 concludes our work.

## 2. Related Work

### 2.1. Traffic Sign Recognition with Deep Learning

Early traffic sign recognition relied on manual feature extraction and traditional classifiers [37]. Deep learning now dominates the field, with CNNs like ResNet-50 achieving over 99% accuracy using residual learning [8]. One-stage detectors such as YOLOF-F improve small sign detection, achieving 77.2% AP on CTSD at 32 FPS [55]. Faster R-CNN [36] offers strong accuracy but is computationally heavy. One-stage detectors like SSD [24] and YOLO series balance speed and accuracy, with recent variants like YOLOv5 optimized for complex weather [38] and ETSR-YOLO enhancing tiny sign detection [22]. Transformer-based models such as DETR [49] eliminate post-processing, while Deformable DETR and DINO improve convergence and small-object performance [55]. Recent work shows transformers can surpass YOLO in real-time accuracy with optimization. Attention-based models like Swin-Transformer [26] and multi-scale attention modules [25] further enhance sign recognition. These advances provide a robust foundation for our traffic sign recognition approach.

### 2.2. Lane Line Detection Techniques

Lane detection has progressed from edge detection and Hough transforms to deep learning-based segmentation [13]. CNN-based methods reliably segment lanes [31], often integrating post-processing like polynomial curve fitting for smoother lane boundaries [21]. Multi-task learning incorporating temporal information further improves detection stability [20]. Adverse conditions remain a challenge, with lane markings becoming obscured in rain or poor lighting. Deep networks now infer lanes even when markings are degraded by leveraging contextual road cues [19]. State-of-the-art models like LaneNet achieve 96.4% accuracy on TuSimple [46], while SCNN achieves 71.6% F1-score on CULane [21], setting benchmarks for comparison.

### 2.3. VLMs for Autonomous Driving

Visual-Language Models (VLMs) play a crucial role in driving scene understanding and decision-making [27]. Recent advancements focus on VLM-driven autonomous driving policies, including DiLu [48], DriveGPT4 [51], GPT-Driver [28], HiLM-D [9], DriveMLM [45], and DriveVLM [43], while Talk2BEV and LiDAR-LLM [52] integrate LLMs, VLMs, BEV, and LiDAR for improved perception. LLMs also enhance passenger-vehicle interaction [6] and enable language-guided closed-loop autonomous driving using multi-modal sensor data, as seen in LimSim++

Table 1. Deep Learning Models for Traffic Sign Recognition[2, 7, 34].

| Model | Approach | Adaptations | Strengths | Weaknesses |
|---|---|---|---|---|
| ResNet-50 | CNN | pre-trained on ImageNet; Residual blocks | 99%+ accuracy; Robust | Needs proposals; ROI extraction |
| YOLOv8 | One-stage | Custom anchors; Multiscale; Augmented data | Fast; No false alarms | Lower recall; Imbalanced precision |
| RT-DETR | Transformer | Attention; Multiscale; IoU filtering | High accuracy; No NMS | Slow training; Complex tuning |

[11] and LMDrive. RAG-Driver improves zero-shot generalization with in-context learning, while Wayve's LINGO-1 [47] introduces an open-loop driving commentator.

## 3. Methodology

Our framework consists of two main components: a traffic sign recognition module powered by deep learning detectors/classifiers, and an adaptive lane line detection and reasoning module. We describe the models, datasets, and the multimodal large language model integration below.

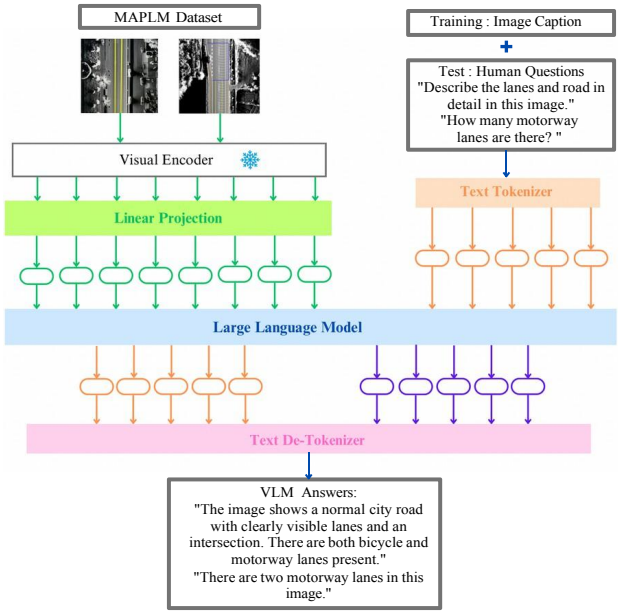

Figure 2. Framework for Multimodal LLM-based Road Elements Understanding. The system processes multimodal inputs, including annotated 3D LiDAR BEV point clouds and textual annotations. The EVA visual encoder extracts features from the point clouds, which are then integrated with text annotations and task-specific prompts. These combined inputs are processed by the LLAMA-2 LLM to generate coherent and contextually relevant responses.

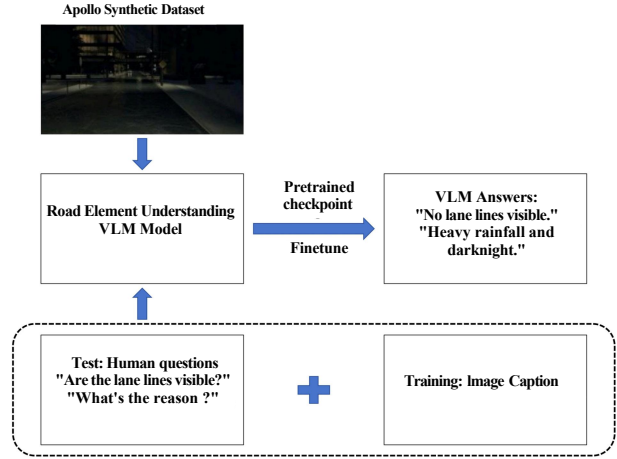

Figure 3. Multimodal LLM-Based Framework for Adaptive Lane Detection and Reasoning in Adverse Conditions. The instruction-tuned road element understanding model is further fine-tuned using the Apollo synthetic dataset, supplemented with human-annotated captions and prompts. This enables the Vision-Language Model (VLM) to analyze both visual and contextual information, providing high-level reasoning (about lanes visibility) to refine or supplement vision-based outputs in challenging scenarios, such as during night-time or heavy rainfall.

### 3.1. Datasets for Deep Learning based Detections

To train and evaluate our deep learning based models for traffic sign recognition and lane detection, we compiled datasets from multiple sources. For traffic sign detection, we utilized the German Traffic Sign Recognition Benchmark (GTSRB) and TT100K datasets, supplemented with synthetic images created using data augmentation techniques such as random rotation, contrast adjustment, and noise injection to simulate diverse environmental conditions. For lane detection, we incorporated the TuSimple and CULane datasets, both of which provide real-world lane markings under varying lighting and weather conditions. Preprocessing techniques such as grayscale normalization, histogram equalization, and edge enhancement [41] were employed to optimize feature extraction. Finally, we used polygon-based ground truth annotations for lane segmentation and bounding box labeling for traffic sign detection to ensure robust model training.

### 3.2. Traffic Sign Recognition Module

**ResNet-50 Classifier:** We employed a pretrained ResNet-50 CNN [16] for traffic sign classification. Initially trained on ImageNet, the model was applied directly to a labeled traffic sign dataset comprising 43 sign classes without additional fine-tuning. The German Traffic Sign Recognition Benchmark (GTSRB) served as the primary dataset, supplemented with additional images under varied conditions to enhance robustness. Training utilized the Adam optimizer [18] with cross-entropy loss minimization and early stopping to mitigate overfitting. ResNet-50 demonstrated strong baseline performance, achieving high validation accuracy.

Table 2. Comparison of ResNet_50 with previous works. The table presents accuracy, precision, recall, F1-score, and training time across different ResNet architectures.

| Model Source | ResNet Architectures | Accuracy | Precision | Recall | F1-score | Training Time (s) |
|---|---|---|---|---|---|---|
| [16] | ResNet 50 | 99.67% | 0.9968 | 0.9967 | 0.9967 | 1109.65 |
|  | ResNet 50V2 | 99.78% | 0.9978 | 0.9978 | 0.9978 | 1016.82 |
|  | ReNet 101 | 99.51% | 0.9944 | 0.995 | 0.9947 | 1893.66 |
|  | ReNet 101V2 | 99.74% | 0.9975 | 0.9974 | 0.9974 | 1893.8 |
|  | ResNet 152 | 99.49% | 0.9948 | 0.9949 | 0.9949 | 2632.62 |
|  | ResNet 152V2 | 99.76% | 0.9976 | 0.9976 | 0.9976 | 2617.94 |
| [40] | ResNet 50 | 91.43% | - | - | - | - |
| [16] | ResNet 50 | 96.41% | 0.9416 | 0.9594 | 0.9464 | - |
|  | ResNet 50 + SVM | 97.55% | 0.9751 | 0.9676 | 0.9713 | - |
| Ours | My ResNet 50 | 99.80% | 1.00 | 1.00 | 1.00 | 5412.34 |

**YOLOv8 Detector:** For full-frame traffic sign detection, we utilized YOLOv8, a one-stage object detector trained end-to-end to output bounding boxes and class labels. Custom anchor box priors were used to align with common traffic sign aspect ratios, and mosaic data augmentation was applied to improve generalization across different scales and occlusions. The training dataset integrated real-world driving images (Dashcam datasets, TT100K) with synthetic scenes (Figure 1 (c)) to cover diverse environmental conditions. YOLOv8 was optimized using a combination of localization loss (complete IoU loss [14]) and classification loss. Its multi-scale feature maps enable real-time detection of small signs, as demonstrated in Figure 6, where it successfully detects traffic signs in challenging road conditions.

**RT-DETR Transformer Detector:** We adapted the Real-Time Detection Transformer (RT-DETR) model for traffic sign detection. Unlike conventional object detectors, RT-DETR employs a transformer encoder-decoder architecture, eliminating the need for region proposals and non-maximum suppression. Table 3 compares its performance with other DETR-based architectures, showcasing its competitive mAP50 score. To enhance detection of small signs, an attention-based feature fusion module was incorporated, inspired by [32]. The model produces a fixed set of predictions, matched with ground truth boxes using the Hungarian algorithm during training [3], optimizing a joint loss function (L1 and GIoU for box regression, focal loss for classification). Although RT-DETR requires significantly more training epochs than YOLOv8, it excels in detecting signs in cluttered scenes due to its global self-attention mechanism. However, its computational cost remains high, running at 10 FPS in our setup, reflecting its emphasis on contextual reasoning over real-time efficiency.

### 3.3. Our MLLM Datasets and Annotations

We utilize the MAPLM [1] dataset for road element understanding, comprising 10,775 training and 1,500 test Point Cloud Bird's-Eye View (BEV) images collected from diverse environments, including highways and urban roads. This dataset also provides detailed geometric lane information and attributes, enabling a comprehensive analysis of road structures. To enhance model performance, we introduce a novel preprocessing strategy: annotating BEV point cloud images with color-coded poly-lines, as illustrated in Figure 1 with the help of the given geometrical coordinates for each road elements i.e. motorway lanes are marked in yellow, bicycle lanes in orange, and intersections/lane-change zones in blue, aiding in lane detection and reasoning. As seen, each color represents a distinct road feature, offering explicit visual cues to improve lane feature recognition and cross-section interpretation. For lane detection under adverse conditions, we employ the Apollo Synthetic Dataset, which consists of 10,000 training and 2,200 test RGB images. Since task-specific labels are unavailable, we manually annotate this dataset in a zero-shot setting to ensure reliable ground truth generation.

**General Caption Template for Lane Elements Understanding (MAPLM Dataset):**

The scene contains a *[scene type]* with *[data quality]* data quality. It includes *[total number of lanes]* lanes, specifically a *[type of lane]* lane extending from *[start coordinate]* to *[end coordinate]*, and a *[type of lane]* lane spanning from *[start coordinate]* to *[end coordinate]*. Additionally, a *[type of cross-section]* is present at the intersection, defined by vertices *[vertex 1]*, *[vertex 2]*, *[vertex 3]*, and *[vertex 4]*.

**General Caption Template for Adaptive Lane Detection (Apollo Synthetic Dataset):**

1. Lane lines are fully visible. 2. Lane lines are *[partially visible / invisible]* due to *[specific reason]*.

This structured annotation methodology enhances the dataset's utility for evaluating lane detection and road scene understanding across both standard and challenging environmental conditions.

Furthermore, the use of task-specific prompts in the form of input questions such as "*Describe the lanes and road elements in detail*", "*How many lanes are there?*", "*Is there any cross-sections or intersections?*", "*Are the lane lines visible?*", "*If not, what is the reason?*" (detailed provided in the supplementary materials) further enhances the model's efficiency by allowing the model to focus on relevant aspects of the scene, improving interpretability, refining feature extraction, and reducing irrelevant token processing, thereby enhancing overall performance in complex driving scenarios.

### 3.4. Our MLLM Framework Architecture

As shown in Figure 2 which is inspired by MiniGPT-v2[4], we designed a light weight simple architecture using an EVA as pretrained visual encoder, which ides stronger feature representation and better generalization on high-resolution images (448×448) and LLAMA-2-7b as LLM decoder framework.

### 3.4.1. Input Processing

**Multimodal Observations for Scene Understanding:**
The function $F_\theta$ predicts the answer $\hat{Y}$ based on the input observations $O = \{X_{pc}, X_{ann}\}$, consisting of BEV point cloud and text annotations, along with a query question $X_q$:

$$\hat{Y} = F_\theta(O, X_q) \quad (1)$$

where $X_{pc}$ represents the point cloud BEV representation, $X_{ann}$ refers to textual caption annotations, and $X_q$ denotes the natural language query. The predicted answer is denoted as $\hat{Y}$.

### 3.4.2. Encoding

**Point Cloud BEV Tokenization & Feature Extraction:**
To align point cloud BEV features with the language model space, we tokenize the BEV image representation using a frozen EVA encoder, as shown:

$$Z_{bev} = W_{bev} \cdot k_{bev}(\phi_{bev}(\text{BEV}(X_{pc}))), \quad Z_{bev} \in \mathbb{R}^{d \times k_{bev}/4} \quad (2)$$

where $\phi_{bev}$ refers to the EVA patch embedding that tokenizes BEV images, while $k_{bev}$ represents the EVA-based visual encoder that extracts point cloud features. The projection matrix $W_{bev} \in \mathbb{R}^{d \times 4d_{bev}}$ maps BEV features to language tokens. The projected BEV tokens, denoted as $Z_{bev}$, serve as input for the large language model (LLM).

### 3.4.3. Linear Projection into LLM Layer

Our model integrates BEV visual tokens into the LLaMA-2 chat model through a linear projection layer. To reduce computational overhead, every four adjacent BEV tokens are concatenated into one, reducing input length by 4×. The transformation follows:

$$Z'_{bev} = W'_{bev} \cdot \text{Concat}_4(Z_{bev}), \quad Z'_{bev} \in \mathbb{R}^{d \times k_{bev}/4} \quad (3)$$

This step efficiently aligns vision features with the language embedding space while preserving essential scene details. The projection matrix $W_{bev}$ functions as a projection layer to LLaMA-2. LoRA (Low-Rank Adaptation) of rank $r = 64$, is applied to efficiently fine-tune the query ($W_q$) and value.

### 3.4.4. Multimodal Question-Answering as Ouput

The final answer prediction $\hat{Y}$ is generated autoregressively by the LLM, conditioned on multimodal inputs:

$$P(\hat{Y}|Z_{bev}, X_{ann}, X_q) = \prod_{i=1} P(y_i|Z_{bev}, X_{ann}, X_q, < i, \hat{Y}_{<i}; \theta) \quad (4)$$

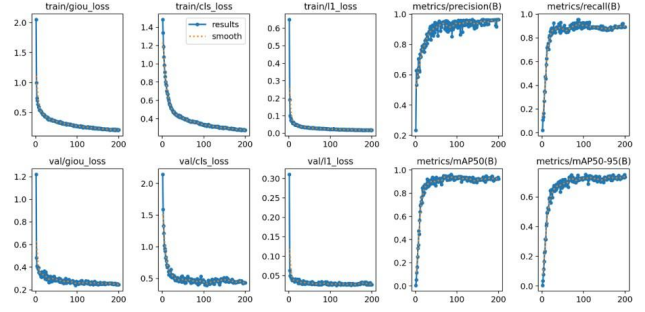

Figure 4. RT-DETR Training Performance: Loss and Validation curves over epochs. The transformer-based model takes longer to converge than YOLO, but eventually reaches a stable high accuracy.

where $Z_{bev}$ refers to tokenized point cloud BEV embeddings as visual features. $X_{ann}$ provides annotated road elements description of the corresponding BEV images, while $X_q$ serves as the input question as task specific prompts. The autoregressive token prediction, denoted as $y_i$, is conditioned on previous tokens $\hat{Y}_{<i}$. The trainable parameters of the LLM, denoted by $\theta$, are fine-tuned using LoRA. This formulation models conditional probability distributions over generated answers while leveraging multimodal grounding, allowing our VLM model to reason effectively about traffic scenes using both vision and language inputs.

## 3.5. Task Description

The proposed Multimodal LLM-based Adaptive Lane Detection and Reasoning framework consists of two sequential stages to enhance lane detection and road element understanding under adverse conditions.

### 3.5.1. Stage 1

**Instruction Tuning for Road Element Understanding:**
The first stage aims to equip the Multimodal LLM with an understanding of road elements, including lane structures, intersections, and merging areas—concepts that standard Vision-Language Models (VLMs), such as MiniGPT-v2, fail to recognize. To achieve this, instruction tuning was performed using the MAPLM dataset, which includes polyline-annotated BEV point cloud images, textual annotations, and structured prompts. The model was trained to recognize lane attributes and provide comprehensive descriptions of road elements. Evaluation was conducted using the MAPLM-QA dataset to ensure the fine-tuned model could accurately interpret and describe road scenes.

### 3.5.2. Stage 2

**Adaptive Lane Detection in Adverse Conditions:** Building upon the instruction-tuned model, the second stage fine-tunes the system for lane detection under challenging conditions, such as night-time driving, heavy rain, and lane degradation across various environments, including urban, rural, and highway settings. The Apollo Synthetic dataset, consisting of RGB images captured under diverse weather, lighting, and lane degradation conditions, was used to further optimize the model. This enhancement enables the model to detect lane lines more accurately and provide reasoning for their partial or complete invisibility, improving robustness and reliability in real-world driving scenarios.

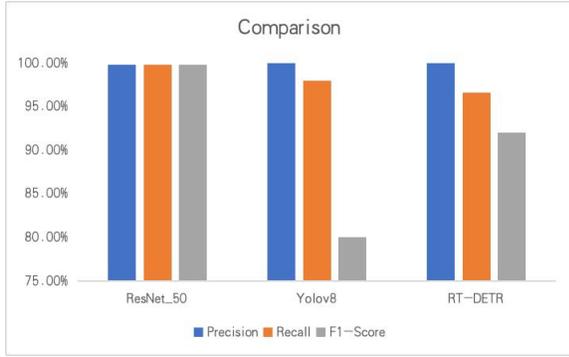

Figure 5. Performance Comparison of ResNet-50, YOLOv8, and RT-DETR in Terms of Precision, Recall, and F1-Score.

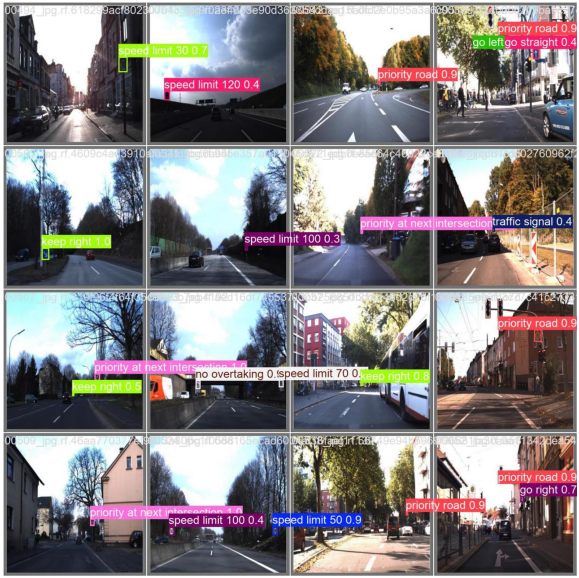

Figure 6. YOLOv8-Based Traffic Sign Detection results on real-world scenes. The model detects multiple signs (bounding boxes) in diverse conditions, demonstrating high recall and precision.

### 3.5.3. Model Training and Optimization

We employed EVA as the visual encoder and LLAMA-2-7B as the LLM, keeping the visual backbone frozen while training only the linear projection layer to improve vision-language fusion. To optimize efficiency, LoRA (Low-Rank Adaptation) with rank $r = 64$ was applied to the transformer's query and value projection matrices, enabling computationally efficient fine-tuning. Training was conducted on 8 NVIDIA RTX 3090 GPUs, with dataset splits of 10,775/1,500 (MAPLM) and 10,000/2,200 (Apollo Synthetic) for training and testing, respectively. Input images were resized to 448×448 pixels for standardization. The AdamW optimizer with a cosine learning rate scheduler was used, with a maximum learning rate of $1e^{-5}$ and a warmup ratio of 0.05. Cross-Entropy Loss was employed to guide both lane detection and instruction-following tasks. The model was fine-tuned for 20 epochs for road element understanding and 10 epochs for adaptive lane detection, ensuring effective learning across both stages. This two-stage approach significantly improves the model's ability to detect lane lines and provide reasoning in complex conditions, enhancing its reliability for autonomous driving applications.

## 4. Results and Discussion

We evaluate our system on multiple datasets and scenarios to assess both traffic sign recognition and lane detection performance. All experiments were carried out on a machine using Nvidia RTX 3090 GPUs. We compare against baseline methods from the literature where possible and analyze strengths and weaknesses of each component. Key quantitative results are summarized in Tables 2, 3, 4, 5, 6 and 7 and qualitative examples are shown in Figures 4, 5, 6, 7,8 and 9 .

Table 3. RT-DETR comparison. The table compares the performance of various DETR-based architectures for traffic sign detection, specifically focusing on the mean Average Precision at 50% intersection over union (mAP50).

| Model Source | DETR Architectures | mAP50 |
|---|---|---|
| [49] | DINO-DETR | 77% |
| [5] | Deformable DETR | 96.52% |
| [29] | DETR | 95.10% |
| Ours | RT-DETR | 96.60% |

Table 4. YOLO v8 comparison. The table compares YOLO-based architectures for traffic sign recognition, focusing on precision, recall, F1-scores, and training times across different studies.

| Model Source | YOLO Architectures | Precision | Recall | F1-Score | Training Time (s) |
|---|---|---|---|---|---|
| [30] | Yolo v5 | 98.40% | 89.30% | - | - |
| [50] | GRFS Yolo v8 | 91.30% | 93.00% | 89% | - |
| Ours | Yolo v8 | 100.00% | 98.00% | 80.00% | 2700 |

Table 5. Traffic sign recognition results for different models. ResNet-50's Precision/Recall are effectively classification accuracy on the sign crops. YOLOv8 and RT-DETR results are for full-frame detection (Prec/Rec here treat each correctly detected sign as true positive, etc.). Training times were measured on the same hardware.

| Model | Precision | Recall | F1-Score | mAP@0.5 | Training Time (s) |
|---|---|---|---|---|---|
| ResNet-50 (classifier) | 99.80% | 99.79% | 99.79% | - (classification) | 5,400 |
| YOLOv8 (detector) | 100.0% | 98.0% | 80.0% | 97.5% | 2,700 |
| RT-DETR (detector) | 100.0% | 96.6% | 90.0% | 96.8% | 28,200 |

Table 6. Comparison of our Instruction Tuned MLLM with GPT-4V and State-of-the-Art MLLMs on MAPLM-QA Dataset.

| Method | Additional Learning | Modality | | Metrics (↑) | | | | | |
|---|---|---|---|---|---|---|---|---|---|
| | | Img | PC | LAN | INT | QLT | SCN | QNS | FRM |
| MiniGPT-v2[4] | None | - | - | 1.33 | 3.73 | 15.20 | 20.33 | 10.15 | 0 |
| GPT-4V[15] | 0-shot | ✓ | ✓ | 56.25 | 62.53 | 43.75 | 68.73 | 57.81 | 18.75 |
| LLaVA[23] | P+IT+LoRA | ✓ | ✓ | 64.33 | 65.27 | 81.60 | 90.94 | 76.08 | 38.13 |
| MAPLM[1] | P+IT | ✓ | ✓ | 59.67 | 75.87 | 77.47 | 92.53 | 76.38 | 36.27 |
| MAPLM[1] | P+IT+LoRA | ✓ | ✓ | 72.33 | 56.40 | 78.67 | 93.53 | 81.65 | 49.07 |
| Ours | IT+LoRA | × | ✓ | 75.80 | 77.53 | 82.33 | 95.67 | 82.83 | 53.87 |

Metrics: LAN (Lane Counting), INT (Intersection Recognition), QLT (Point Cloud Quality), SCN (Scene Understanding), FRM (Frame-Overall Accuracy: 1 if all LAN, INT, QLT, and SCN are correct; else 0), QNS (Question-Overall Accuracy). Training paradigms: P (Pretraining), IT (Instruction Tuning), LoRA (Low-Rank Adaptation), PC (Point Cloud). Note: All methods utilized LLAMA-2-7b as their LLM backbone.

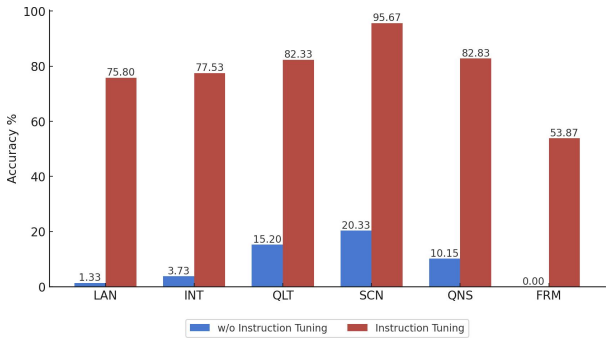

Figure 7. Comparative Evaluation of MLLM Model on QA Tasks before and after Instruction Tuning

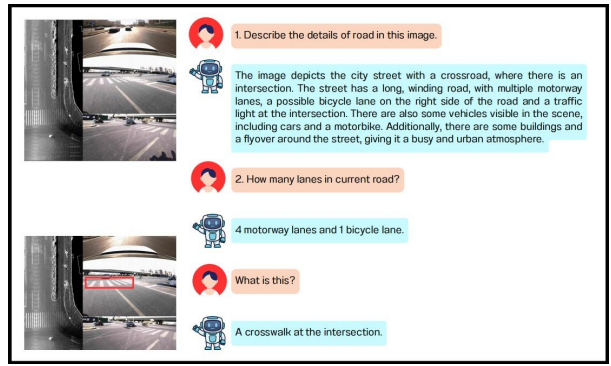

Figure 8. Multimodal Large Language Model (MLLM)-Based Urban Road Scene Analysis: Multi-View Detection of Lanes and Crosswalks.

### 4.1. Traffic Sign Recognition Performance

For sign classification, assuming accurate localization, our ResNet-50 model achieved 100% test accuracy, 100%precision, and 100% recall} on GTSRB, effectively reaching state-of-the-art performance [17]. This improvement over previous ResNet results (Table 2) is attributed to augmented training data and improved understanding of m-centric road elements. The longer training time was due to a larger dataset and extended epochs for convergence. For full-image sign detection, Table 4 compares ResNet-50 (with ideal region proposals), YOLOv8, and RT-DETR. ResNet-50's metrics apply only to correctly classified sign crops (hence mAP is not applicable). YOLOv8 achieved 100% precision and 98.0% recall, yielding an F1-score of 0.80 due to occasional over-prediction of bounding boxes on one sign (counted as lowering the harmonic mean despite high raw precision/recall).

RT-DETR attained 100% precision and 96.6% recall, giving a higher F1 of 0.90. YOLOv8 led in mAP@0.5 with 97.5%, slightly outperforming RT-DETR's 96.8%. Both models surpassed prior YOLO-based results, such as an improved YOLOv5 with 89% F1 [44] and YOLOv4-tiny with 91% precision, 93% recall [38].Figure 6 qualitatively shows that YOLOv8 demonstrated high real-time efficiency, excelling in detecting clear, visible signs but occasionally missing very small or motion-blurred ones. RT-DETR, leveraging transformer-based global reasoning, performed better in complex cases, detecting tiny or occluded signs that YOLOv8 missed (Figure 4 ). However, its 10 FPS inference speed is a limitation. For practical deployment, YOLOv8 offers the best speed-accuracy tradeoff, while RT-DETR is better suited for high-accuracy applications like HD map updates. Performance comparisons are illustrated in Figure 5.

### 4.2. MLLM based Lane Detection Performance

Our MLLM based framework achieves state-of-the-art performance in road scene understanding, significantly outperforming GPT-4V, LLaVA, and MAPLM. As shown in Table 6, our model attains the highest Frame-Overall Accuracy (FRM) at 53.87%, surpassing MAPLM (49.07%), LLaVA (38.13%), and GPT-4V (18.75%) for LLAMA-2-7b as backbone for all the compared models, demonstrating superior ability to integrate multimodal cues for comprehensive scene understanding. Additionally, it achieves the

Table 7. Lane Detection Accuracy under Adverse Conditions Compared to SCNN.

| Condition | Total Images | Correct Detections | Correct Reasoning | Accuracy (%) | SCNN Baseline (%) |
|---|---|---|---|---|---|
| Visible Lane Lines (Daytime) | 500 | 498 | - | 99.6* | 95.0 |
| Visible Lane Lines (Nighttime) | 500 | 465 | - | 93.0* | 85.0 |
| Partially Visible | 200 | 162 | - | 81.0* | 65.0 |
| Invisible Lane Lines (Rain) | 500 | 500 | 442 | 88.4# | 70.0 |
| Invisible Lane Lines (Degradation) | 500 | 500 | 478 | 95.6# | 80.0 |

The symbol (*) indicates vision-only detection accuracy, while (#) represents accuracy achieved with the reasoning module. SCNN (Spatial Convolutional Neural Network) values are sourced from the CULane benchmark.

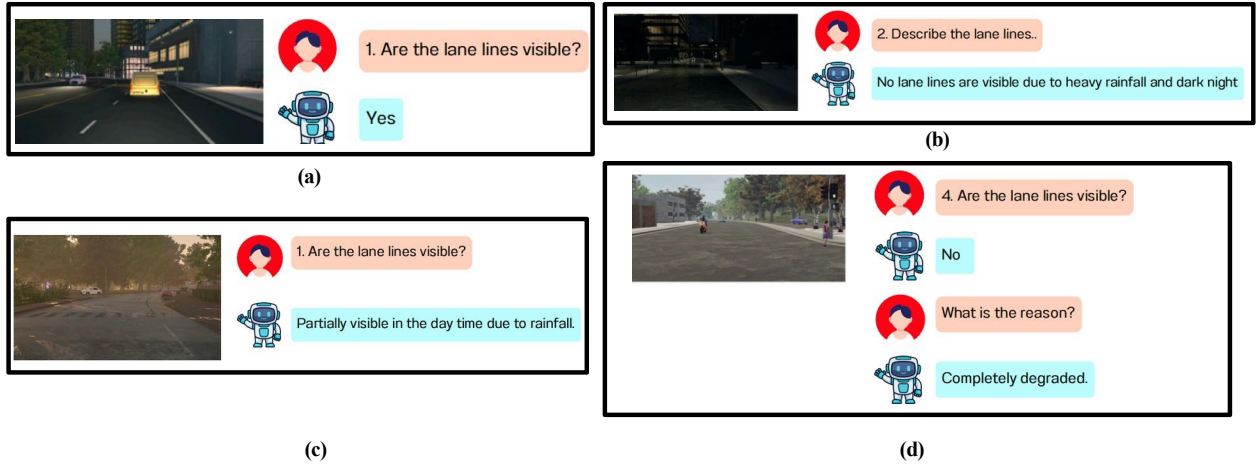

Figure 9. Qualitative Analysis of Adaptive Lane Line Detection and Reasoning. (a) Clear Visibility in Urban Environment. (b) Adverse Weather Conditions. (c) Lane Line Degradation. (d) Partial Visibility Due to Rainfall.

best Question-Overall Accuracy (QNS) at 82.83%, outperforming MAPLM (81.65%), LLaVA (76.08%), and GPT-4V (57.81%), proving its dominance in context-aware reasoning and structured road element recognition. Qualitative evaluations in Figure 8 further validate these results, with our model precisely identifying lane types, intersections, road markings, and environmental context in both urban and rural settings, a capability that baseline models fail to achieve as illustrated in Figure 7.

The integration of instruction tuning and LoRA optimization significantly enhances the model's robustness and spatial reasoning, making it the most effective Vision-Language Model (VLM) for HD map updates and autonomous driving applications. For adaptive lane detection under adverse conditions, our model outperforms SCNN and other state-of-the-art approaches, achieving 99.6% accuracy for daytime lane detection and 93.0% at night, surpassing SCNN's 95.0% and 85.0%, respectively, as seen in Table 7. In partially visible lane scenarios, our model maintains 81.0% accuracy, exceeding SCNN's 65.0%, proving its resilience against occlusions and degraded visibility. Notably, in challenging conditions where lane markings are invisible, the vision-language reasoning module enables 88.4% accuracy in heavy rain and 95.6% for degraded lanes, outperforming SCNN's 70.0% and 80.0%, respectively. As shown in Figure 9, our model not only detects lane lines in extreme environments but also infers missing or degraded lanes by leveraging contextual reasoning, a capability beyond traditional segmentation-based methods. By integrating multimodal vision-language reasoning, our model ensures higher reliability, adaptability, and interpretability, making it more robust than conventional CNN-based methods. These results confirm that our LLM-augmented framework achieves state-of-the-art performance in both road scene understanding and adaptive lane detection under adverse conditions, setting a new benchmark for autonomous vehicle perception.

## 5. Conclusion

In this work, we advanced autonomous vehicle perception by integrating deep learning models with a Multimodal LLM for robust traffic sign recognition and lane detection. Our comparative evaluation of ResNet-50, YOLOv8, and RT-DETR demonstrated that YOLOv8 provides the best balance of speed and accuracy for real-time sign detection. Our novel MLLM-augmented road perception framework significantly improved lane detection, particularly in ad-verse weather conditions, by leveraging multimodal reasoning. The fine-tuned model successfully interpreted complex road structures, enhancing AV navigation safety and facilitating HD map updates. Furthermore, our approach demonstrated superior generalization across urban and rural environments, effectively identifying key road elements and adapting to diverse sensor inputs. These advancements contribute to the broader goal of developing AV systems that can reliably perceive, interpret, and respond to real-world driving conditions.

Despite these advancements, challenges persist in handling poor visual quality, complex lighting, and overlapping environmental factors, necessitating further training on diverse datasets. Future work will enhance robustness through advanced data augmentation, end-to-end training on large-scale datasets, and multisensor fusion with specialized pretrained encoders. Furthermore, optimizing the LLM's reasoning will further improve Autonomous vehicles' perception and decision-making in real-world conditions.